\documentclass[letterpaper]{article} 
\usepackage{aaai23}  
\usepackage{times}  
\usepackage{helvet}  
\usepackage{courier}  
\usepackage[hyphens]{url}  
\usepackage{graphicx} 
\usepackage{natbib}  
\usepackage{caption} 
\usepackage{algorithm}
\usepackage{algorithmic}
\usepackage{booktabs}
\usepackage{graphicx}
\usepackage{amsmath}

\usepackage{newfloat}
\usepackage{listings}
\DeclareCaptionStyle{ruled}{labelfont=normalfont,labelsep=colon,strut=off} 
\floatstyle{ruled}
\newfloat{listing}{tb}{lst}{}
\floatname{listing}{Listing}
\title{Leveraging BERT Language Model for Arabic Long Document Classification}
\author {
    Muhammad AL-Qurishi
}
\affiliations{
    Elm Company\\
    Research Department\\
    Riyadh 12382, Saudi Arabia\\
    mualqurishi@elm.sa
}

\usepackage{bibentry}

\begin{document}

\maketitle

\begin{abstract}
Given the number of Arabic speakers worldwide and the notably large amount of content in the web today in some fields such as law, medicine, or even news, documents of considerable length are produced regularly. Classifying those documents using traditional learning models is often impractical since extended length of the documents increases computational requirements to an unsustainable level. Thus, it is necessary to customize these models specifically for long textual documents. In this paper we propose two simple but effective models to classify long length Arabic documents. We also fine-tune two different models-namely, Longformer and RoBERT, for the same task and compare their results to our models. Both of our models outperform the Longformer and RoBERT in this task over two different datasets.
\end{abstract}

\section{Introduction}
A large portion of textual content that requires automated processing is in the form of long documents. In some domains such as legal or medical, long documents are the standard. This severely restricts the possibilities for practical use of the most advanced Transformer models for text classification and other linguistic tasks~\cite{tay2020long}. For example, models such as BERT~\cite{devlin2018bert} have significantly improved the accuracy of automated NLP tasks, but their usefulness is limited to relatively short text sequences~\cite{ding2020cogltx} due to the fact that their complexity increases geometrically. Modifying BERT in such a way to disassociate sequence length from computing complexity would remove this obstacle and bring immediate benefits to numerous fields such as education, science, and business~\cite{wagh2021comparative}. Innovative approaches that leverage the greatest advantages of Transformers while offsetting their major shortcomings are needed at this stage of development, as they could lead to full maturation of a concept that has been demonstrated to be impressively successful with semantic tasks.

There have been numerous attempts to improve the performance and efficiency of BERT with long documents, using a wide variety of approaches. Some of the proposed solutions are based on the sliding window paradigm~\cite{wang2019multi, pappagari2019hierarchical}. The downside of this class of solutions is their inability to track long-range dependencies in the text which weakens their analytic insights. Another group of works aim to simplify the architecture of Transformers and decrease complexity as result~\cite{sukhbaatar2019adaptive,tay2020efficient,rae2019compressive}. So far, none of these attempts could match the same level of performance that BERT achieves with short text. Reusing previously completed steps is another strategy for adapting Transformers for longer text~\cite{dai2019transformer} as a prominent example. Longformer model proposed by~\cite{beltagy2020longformer} may be the most promising solution for the problem of using Transformers with long text, and it combines local and global attention to improve efficiency. The issue remains open, and new suggestions for the best method of long document processing are still being made on a regular basis.

In this paper we present two BERT-based language models and fine-tune two others for Arabic long document classification. The first language model consists of four main layers: sentence segmentation layer, BERT layer, a linear classification layer, then the sentence grouping layer with respect to each document, and finally the softmax layer. In this model, we segmented the document into meaningful sentences and then fed these sentences into BERT model along with their document ID. The second model has the same idea of dividing the document into sentences, but instead we hypothesize that a majority of semantically important information is concentrated within specific sentences inside of a longer text, making it unnecessary to check for connections between all words in a document. Instead, we used BERT-based similarity match algorithm that can recognize high-relevance sentences and pass them as input to the BERT-base model that can complete the desired classification task. Both of those models are based on BERT architecture, and require supervised training for best performance. Input text is divided into sentences that don’t exceed the maximum length that BERT can accurately process (512 tokens).

In addition, we have fine-tuned two well-known language models for long documents classification task, which are the Longformer~\cite{beltagy2020longformer} and Recurrent over BERT~\cite{pappagari2019hierarchical}\footnote{https://github.com/helmy-elrais/RoBERT\_Recurrence\_over\_BERT}. Before the fintuning process, these two models have been modified to be suitable for the Arabic language. We compared the proposed models against the Longfroemer and RoBERT using two different Arabic datasets. The proposed language models were evaluated against two models, Longfarmer and RoBERT, using two datasets. The first dataset was collected from the Mawdoo3 website\footnote{www.mawdoo3.com} and the second dataset was from previous related work~\cite{abbas2005comparison}. The results showed that the first language model based on sentences aggregating after classifying them is the best among all models on news data with a macro F1-score equal to 98\%. While this model achieved a comparative result with the Longformer in the second Mawdoo3 dataset that contains 22 classes.

\begin{figure*}[ht!]
	\centering
	\includegraphics[width=0.65\linewidth]{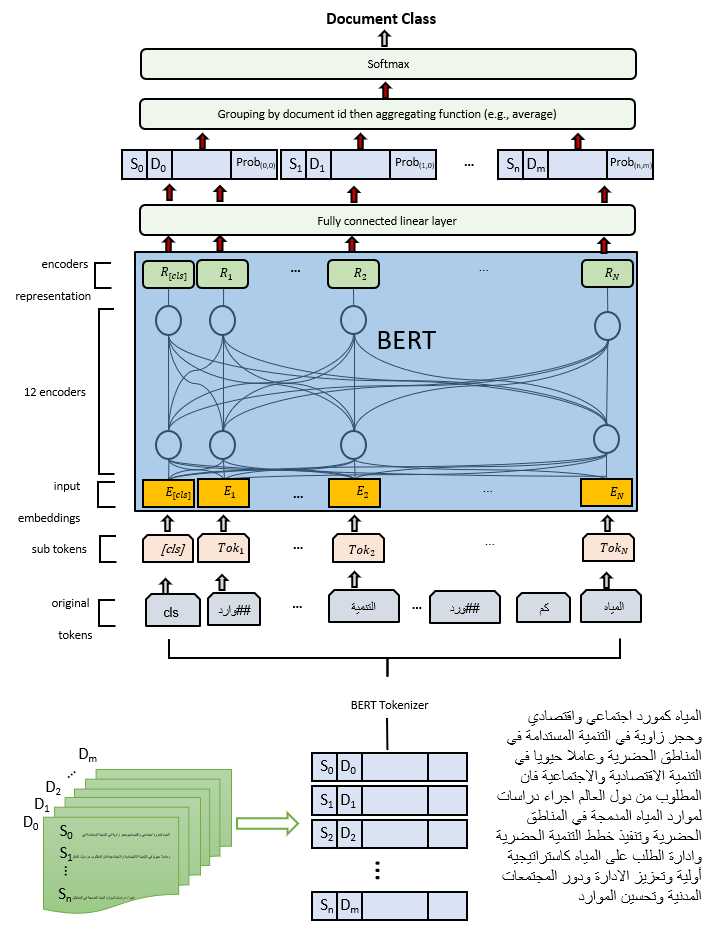}
	\caption{Proposed Model Architecture for Long Document Classification}
	\label{fig:arch2}
\end{figure*}

\section{Related Works}
Most of the recent works addressing the problem of long document classification start from similar principles common to all deep learning methods. They also diverge in many aspects, as the authors explore different avenues for leveraging the power of the learning algorithms and overcoming the most significant obstacles~\cite{dai2022revisiting}. Since the authors are essentially attempting to solve the same problem, namely how to maintain high accuracy of semantic predictions while keeping the computing demands reasonable, it would be fair to describe the papers as belonging to the same family despite the considerable differences in approach.

In terms of methodological choices, practically all works from this group acknowledge the unmatched power of the attention mechanism for analyzing semantic relationships, and incorporate it in some way into the proposed architecture. There is a division between works that mostly (or completely) embrace an existing architecture and perform only minor operations such as fine-tuning or knowledge transfer in order to reduce the computational demands~\cite{adhikari2019docbert,sun2019fine}. On a different end of the spectrum, there are works that propose innovative hybrid solutions in which the attention mechanism and/or Transformer architecture are combined with elements of different deep learning paradigms, such as RNNs and CNNs. In particular, a common strategy is to adopt a hierarchical structure for the overall solution and use the attention mechanism only in a limited role, thus avoiding the exponential growth of complexity~\cite{huang2021hierarchical,si2021hierarchical}.

The aforementioned methodological differences stem largely from the expectations for each paper, which range from proving a theoretical point to attempting to develop specialized model for long document classification. Works with a narrower scope tend to stay closer to the original BERT model design~\cite{beltagy2020longformer}, while more ambitious efforts that aim to create new tools are more inclined to experiment with previously untested combinations of elements. In some papers, the scope of intended applications is limited to long documents from a certain domain (i.e. medical)~\cite{si2021hierarchical}, while others are approaching the problem in more general terms. Finally, there is an important distinction between works that aim for greater accuracy, and those that primarily attempt to improve computational efficiency and shorten the inference time~\cite{park2022efficient}.

It’s a fair assessment that practically all works from this group are grappling with the same problem – the tendency of attention-based models to become prohibitively complex as the length of the analyzed text is increased. In response, the authors tried a variety of ideas that rely on vastly different mechanisms to decrease complexity. From fine-tuning and knowledge distillation to introduction of hierarchical architectures and restrictive elements such as fixed-length sliding window~\cite{beltagy2020longformer, wang2020learning}, the proposed techniques are quite innovative and typically leverage some known properties of deep learning models to affect how the attention mechanism performs in a particular deployment. The diversity of ideas found in those papers illustrates that researchers are currently casting a wide net and searching for unconventional answers to a difficult problem, without a single dominant strategy. On the other hand, hybrid approaches hold a lot of promise and they combine some proven elements from different methodologies into new, potentially more optimal configurations~\cite{huang2021hierarchical,he2019long}.

Evaluation of the proposed changes to established algorithms is crucially important, and all of the reviewed works include some form of empirical confirmation of their premises. While the numbers seemingly validate that the proposed solutions achieve state-of-the-art results under the best possible conditions, those findings are self-reported and may often be too optimistic. All of the papers are interested in document classification tasks and use it to evaluate their solutions, but datasets used for testing may not be the same in terms of size, diversity, and content. When directly comparing different solutions, it’s extremely important to keep in mind the particulars of the evaluation protocols. Studies aiming to provide evaluations with independently administered comparative testing of several different BERT-like algorithms for document classification are slowly emerging and reporting some interesting findings that often diverge from self-assessed results~\cite{wagh2021comparative, dai2022revisiting, park2022efficient}. Still, there are no widely accepted evaluation standards and every comparison suffers from ‘apples-to-oranges’ problem up to an extent.

When it comes to practical use of the proposed solutions, there is a general lack of field data and even discussions of use cases are rare. This is understandable considering the main focus is on discovering more efficient methods, but without real world testing it’s difficult to predict whether any of the solutions can deliver similar results to their reported findings. Some works may be directed as specific niches such as legal or medical, but even in this case little attention is paid to practicalities associated with real world application. This weakness may reflect the current state of the field, which is highly experimental and mostly built on data collected in a controlled environment.

\section{Data}
\label{datSec}
Experimental parts of our study are conducted using two different datasets, with the choice of the datasets based on the domain of research which is long length Arabic documents. The datasets are vastly different in terms of size and diversity of classes.
\subsection{Mawdoo3 Dataset}
\begin{figure*}[ht!]
	\centering
	\includegraphics[width=0.5\linewidth]{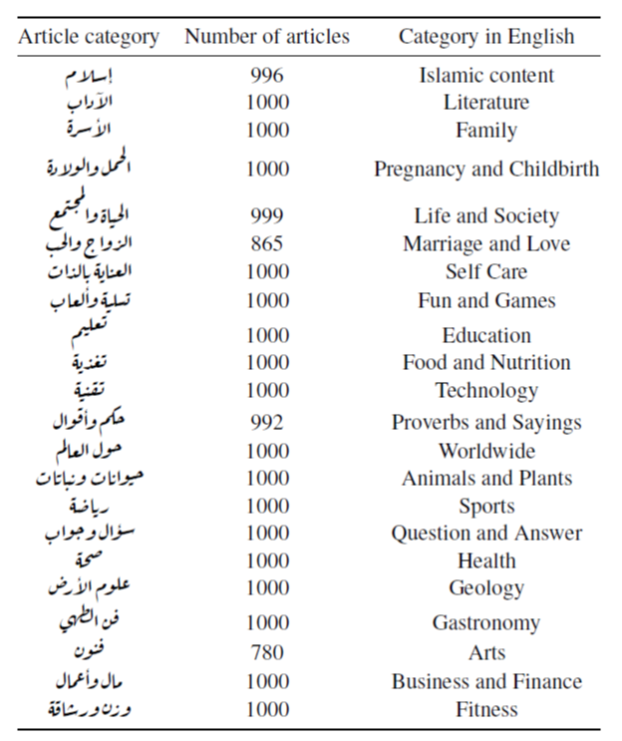}
	\caption{Mawdo3 Dataset that contains 22 class.We selected almost 1000 article under each class.}
	\label{tabMaw}
\end{figure*}

The first dataset was scraped from Mawdoo3 which is the largest Arabic content  website\footnote{https://mawdoo3.com/}. The number of classes from mawdoo3 is 22 class and each category  contains between 700 to 12K articles. We have selected almost one thousand of long articles from each category as presented in Figure~\ref{tabMaw}. 

\subsection{Arabic News Dataset}
\begin{figure*}[ht!]
	\centering
	\includegraphics[width=0.5\linewidth]{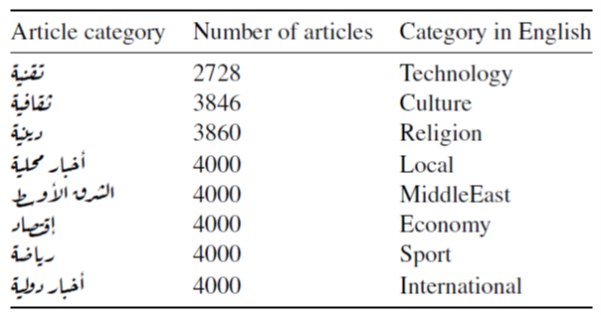}
	\caption{Arabic News Dataset we choose almost 4000 articles from each category.}
	\label{tabNews}
\end{figure*}

The second dataset was about news articles and we downloaded them from different sources~\cite{chouigui2021arabic,abbas2005comparison,abbas2006topic}. These data have almost the same 8 categories so we merged them together and the resulted dataset is described in Figure~\ref{tabNews}. We have selected almost four thousands of long articles from each class.

\section{Models}
In this section we introduce two BERT-based language models. Both of those models are based on BERT architecture, and require supervised training for best performance. Input text is divided into sentences that don’t exceed the maximum length that BERT can accurately process (512 tokens). We also fine-tune two others for Arabic long document classification. the following sections explain that in details.
\subsection{BERT-based Sentence Aggregation}
We propose a simple but effective model to do a long document classification task. Our proposed model consist of multiple layers as shown in figure~\ref{fig:arch2}; namely, sentence segmentation layer, BERT layer, a linear classification layer, then the sentence grouping layer with respect to each document, and finally the softmax layer. The first layer is to make a segmentation of sentences from the long text, taking into account the structure of the sentence in Arabic language. So that the sentence does not lose its meaning or break. The second layer is the BERT tokenizer followed by the embedding representation layer. Since we are using BERT base model named Arabert-V2~\cite{antoun2020arabert}, this layer consists of 12-layer stacked encoders that receive the embedding inputs and process it and send to the an MLP layer.\\
We train the model on all the sentences and each sentence is considered as document. The training outputs are the classification probability for each class as well as the sentence ID and orginal document ID. We make a grouping of text sentences with the probabilities of each category in each sentence, and in the end we aggregate all sentence in the category with the highest probability with respect to the document ID.
\subsection{BERT-based Key Sentences Model}
This model has the same idea of dividing the document into sentences, but instead we hypothesize that a majority of semantically important information is concentrated within specific sentences inside of a longer text, making it unnecessary to check for connections between all words in a document. Instead, we used BERT-based similarity match algorithm that can recognize high-relevance sentences and pass them as input to the BERT model that can complete the desired classification task. The high-relevance sentences were selected by applying a maximal marginal relevance (MMR)~\cite{carbonell1998use}
similarity algorithm as shown in equation~\ref{eq1}. The length of the sentences is between 30 to 150 tokens. 
\begin{multline}
	\label{eq1}
	MMR = argmax_{D_{i}\in X}[\lambda Sim_{1}(D_i,S)-(1-\lambda)\\
	\max_{D{j}\in C} Sim_{2}(D_{i},D_{j})]
\end{multline}
Where $S$ is the sentence vector and $D_i$ is the document vector related to $S$. $X$ is a subset of documents in our dataset we already selected and $\lambda$ is a constant in range of [0–1], for diversification of results. The $Sim_1$ and $Sim_2$ are the similarity function which can be replaced by cosine, euclidean, Jacard and any other distance similarity measures. In our model we have used the proper cosine similarity that explained by equation~\ref{eq2} \footnote{XCS224 Mod2 lecture by Prof. Pott}.
\begin{equation}\label{eq2}
	\frac{cos^{-1}(\frac{\sum_{i=1}^{n}u_i \times v_i}{||u||_2 \times ||v||_2})}{\pi}
\end{equation}
\subsection{Fine-tuned Models}
In this part, we reproduced and fine-tuned two of the important research works in the literature for processing long texts, which are explained in~\cite{pappagari2019hierarchical, beltagy2020longformer}. We have trained and fine-tuned them to classify Arabic long length documents using the datasets mentioned in Tables~\ref{tabMaw} and~\ref{tabNews}.
\subsection{Longformer}
The Longformer~\cite{beltagy2020longformer} was proposed to reduce the complexity of the self-attention matrix. This can be done by making the matrix sparser through the introduction of attention pattern with specified locations that need to be prioritized. By using a sliding window with a fixed length, the model doesn’t enter exponential progression and instead scales linearly with input sequence length. Additional gains can be achieved by dilating the sliding window, which frees up some attention heads to process the overall semantic context while non-dilated heads remain focused on local tokens.\\
However, the implemented restrictions interfere with the model’s ability to be trained for specific tasks, which required the addition of global attention to the model. Linear projections are used to calculate the attention scores, and in this work an extra set of projections related to global attention are used to make training more reliable. The resulting linguistic model has an impressive capacity for contextual analysis, but expends far less computational resources when used with long-form documents than traditional BERT and other Transformer architectures. Nonetheless, Longformer was trained for autoregressive modeling with left-to-right word sequence and train it with Arabic needed some preprocessing. We converted the base model of Arabert-V2  into a Longformer then we fine-tuned the output model for our Arabic long document classification task. 
\subsection{RoBERT}
In this model the authors~\cite{pappagari2019hierarchical} are looking into possible ways to extend the usefulness of the BERT linguistic model to text samples longer than a few hundred words. To do this, they introduce an extension to the fine-tuning procedure and separate the input into smaller chunks. After those chunks are processed by the base BERT model, they are passed through another Transformer or a single recurrent layer before a classification decision is made in the softmax layer. Those variations were named RoBERT (Recurrence over BERT) and ToBERT (Transformer over BERT), collectively described as Hierarchical Transformers because they maintain the hierarchical structure of representations both on the level of extracted segments and the whole document.
Those models were found to converge very quickly when trained on a narrowly focused dataset and to perform better than the original BERT with long text sequences. Suitability of those derivative models was examined for different tasks, including topic identification and satisfaction prediction during a customer call, which are possible real world applications. Unlike the Longformer, with RoBERT the fine-tuning process was straightforward because it was a BERT-Based model.
\section{Experimental approach}
In our work we aim to find a balance between model accuracy on classification task performed over long text sequences and computational simplicity. Therefor we tried to utilize the base version of BERT which have less memory size of 500MB and faster prediction process where the length of the embedding is 768. We used Google Colab pro to train and fine-tune our models. In terms of accuracy, we use standard metrics to track all of those qualities for the tested models. Macro F1 score is used as a general measure of accurate prediction on all comparisons , as it provides a basis for comparison of results between studies.\\
Several hyperparameters have been setup to fine-tune the experimented models. Our proposed classification solutions were tested using two collection of documents mentioned in Sec.~\ref{datSec} where 80\% of the dataset was used to train the model and 10\% as a validation set and 10\% utilized for conducting the tests. Table~\ref{tab1} shows the general parameters used in the training and fine-tuning processes.

\begin{table}[!ht]
	\centering
	\caption{Hyperparameters used in the training and fine-tuning processes}
	\label{tab1}
	\resizebox{\columnwidth}{!}{%
		\begin{tabular}{@{}ll@{}}
			\toprule
			Parameter   Name                                          & Value  \\ \midrule
			number of   epochs                                        & 5      \\
			maximum sequence length aggregation and similarity models & 128    \\
			maximum sequence length longFormer model                & 1024   \\
			maximum sequence length truncation , RoBERT models      & 1024    \\
			mawdoo3 data number of training steps:                  & 107466 \\
			adam epsilon                                            & 1e-8   \\
			train batch   size                                        & 64     \\
			valid batch   size                                        & 128     \\
			epochs                                                    & 20     \\
			learning   rate                                           & 5e-5   \\
			warmup   ratio                                            & 0.1    \\
			max grad   norm                                           & 1.0    \\
			accumulation   steps                                      & 1      \\ \bottomrule
		\end{tabular}%
	}
\end{table}

\section{Results Discussion and Analysis}
The evaluation was conducted using standardized hyper parameters such as batch size and sequence length and others as shown in Table~\ref{tab1}, with two different datasets suitable for Arabic long document classification task as described in Sec~\ref{datSec}. We will try to report the results and analyze them according to each data set separately.
\subsection{Mawdoo3 Dataset}
All models were empirically evaluated on long document classification task. We compared our proposed models with Longformer as well as with the RoBERT on Mawdoo3 dataset. The results were very close between the two proposed solutions and the Longformer, with a very slight superiority to the language model based on extracting key sentences using MMR method with macro F1 score equal to 83\%. While Robert performed very poorly on Mawdoo3 dataset with macro F1 score of 21\%. The overall results of all models in the long document classification task are explained in Table~\ref{tabMawRes}. We can say that this results support our hypothesis of identify the most relevant parts of the text. The resulting solution retains the ability to capture relationships between distant tokens, but doesn’t have to actively back-propagate all of them and instead focuses only on key sentences. Because of this, the model avoids geometric progression of complexity and continues to be efficient with much longer texts than the original BERT is able to. It is worth noting that we have pre-processed and removed the information at the beginning of each article in this dataset because that the parts of the document containing easily identifiable indicators of the class.
\begin{table*}[!ht]
	\centering
	\caption{Overall results of all models in the long text classification task on Mawdoo3 Dataset}
	\label{tabMawRes}
	\resizebox{\textwidth}{!}{%
		\begin{tabular}{@{}lllll@{}}
			\toprule
			Model     & Macro F1 & Macro Precision & Macro Recall & Accuracy  \\ \midrule
			Our Model Aggregating & 0.82187  & 0.82338         & 0.83049      & 0.83083   \\
			Our Model-MMR   & \textbf{0.82732} & \textbf{0.83162}  & \textbf{0.83555} &\textbf{ 0.83522}   \\ 
			Longformer & 0.82347 & 0.82497 & 0.83263 & 0.83291 \\
			RoBERT                & 0.21157 & 0.19309        & 0.29507    & 0.36461\\ \midrule
		\end{tabular}%
	}
\end{table*}

\subsection{Arabic News Dataset}
The results of the experiment were completely different with the Arabic news dataset. All models performed very well, and in this experiment, the first model outperformed the rest with macro F1 score of 98.4\% which revealed that additional modification can have a positive impact on model performance, but it’s important which dataset is used. It was discovered that classifying each sentence is better than classifying the whole sequence, which could even increase performance when working with short sentences. However, both Longformer and our second model with MMR are still performing very well with macro F1 score of 96\% and 96.2\%, respectfully. Whereas RoBERT model has macro F1 score of 74.4\%.  The overall results of all models in the long document classification task are described in Table~\ref{tabANDRes}.
\begin{table*}[!ht]
	\centering
	\caption{Overall results of all models in the long text classification task on Arabic News Dataset}
	\label{tabANDRes}
	\resizebox{\textwidth}{!}{%
		\begin{tabular}{@{}lllll@{}}
			\toprule
			Model     & Macro F1 & Macro Precision & Macro Recall & Accuracy  \\ \midrule
			Our Model Aggregating & \textbf{0.98411}  & \textbf{0.98591 } & \textbf{0.98264}  & \textbf{0.98434}   \\
			Our Model-MMR   & 0.96217 & 0.96240  & 0.96206 &0.96263   \\ 
			Longformer & 0.95908 & 0.95956 & 0.95880 & 0.95961 \\
			RoBERT                & 0.73062 & 0.75382        & 0.75124    & 0.75142\\ \midrule
		\end{tabular}%
	}
\end{table*}

\section{Conclusion}
Unmatched flexibility of BERT is one of the main reasons for its rapid acceptance as state-of-the-art language model. With additional algorithm and some modifications and fine-tuning, the model can be adjusted for certain topics or tasks and its accuracy pushed to even higher level. This work explores this possibility in detail, taking long text classification as the target task and searching for the best parameters for this type of usage. In particular, different possibilities for supervised pre-training and fine-tuning were examined on two different datasets. Through detailed experimentation, we were able to identify the most optimal procedures that enable BERT to be more accurate with our particular downstream task. While the value of the proposed training and tuning actions was confirmed only for text classification, it stands to reason that analogous procedures could prove to be useful for other linguistic tasks as well. Finally, we want to denote that we did not explore all hyperparameters which can be a future work to have along with trying another language models such as Roberta and Electra.

\bibliography{aaai23}

\begin{thebibliography}{24}
\providecommand{\natexlab}[1]{#1}

\bibitem[{Abbas and Berkani(2006)}]{abbas2006topic}
Abbas, M.; and Berkani, D. 2006.
\newblock Topic identification by statistical methods for arabic language.
\newblock \emph{WSEAS Transactions on Computers}, 5(9): 1908--1913.

\bibitem[{Abbas and Smaili(2005)}]{abbas2005comparison}
Abbas, M.; and Smaili, K. 2005.
\newblock Comparison of topic identification methods for arabic language.
\newblock In \emph{Proceedings of International Conference on Recent Advances
  in Natural Language Processing, RANLP}, 14--17.

\bibitem[{Adhikari et~al.(2019)Adhikari, Ram, Tang, and
  Lin}]{adhikari2019docbert}
Adhikari, A.; Ram, A.; Tang, R.; and Lin, J. 2019.
\newblock Docbert: Bert for document classification.
\newblock \emph{arXiv preprint arXiv:1904.08398}.

\bibitem[{Antoun, Baly, and Hajj(2020)}]{antoun2020arabert}
Antoun, W.; Baly, F.; and Hajj, H. 2020.
\newblock Arabert: Transformer-based model for arabic language understanding.
\newblock \emph{arXiv preprint arXiv:2003.00104}.

\bibitem[{Beltagy, Peters, and Cohan(2020)}]{beltagy2020longformer}
Beltagy, I.; Peters, M.~E.; and Cohan, A. 2020.
\newblock Longformer: The long-document transformer.
\newblock \emph{arXiv preprint arXiv:2004.05150}.

\bibitem[{Carbonell and Goldstein(1998)}]{carbonell1998use}
Carbonell, J.; and Goldstein, J. 1998.
\newblock The use of MMR, diversity-based reranking for reordering documents
  and producing summaries.
\newblock In \emph{Proceedings of the 21st annual international ACM SIGIR
  conference on Research and development in information retrieval}, 335--336.

\bibitem[{Chouigui, Ben~Khiroun, and Elayeb(2021)}]{chouigui2021arabic}
Chouigui, A.; Ben~Khiroun, O.; and Elayeb, B. 2021.
\newblock An arabic multi-source news corpus: experimenting on single-document
  extractive summarization.
\newblock \emph{Arabian Journal for Science and Engineering}, 46(4):
  3925--3938.

\bibitem[{Dai et~al.(2022)Dai, Chalkidis, Darkner, and
  Elliott}]{dai2022revisiting}
Dai, X.; Chalkidis, I.; Darkner, S.; and Elliott, D. 2022.
\newblock Revisiting Transformer-based Models for Long Document Classification.
\newblock \emph{arXiv preprint arXiv:2204.06683}.

\bibitem[{Dai et~al.(2019)Dai, Yang, Yang, Carbonell, Le, and
  Salakhutdinov}]{dai2019transformer}
Dai, Z.; Yang, Z.; Yang, Y.; Carbonell, J.; Le, Q.~V.; and Salakhutdinov, R.
  2019.
\newblock Transformer-xl: Attentive language models beyond a fixed-length
  context.
\newblock \emph{arXiv preprint arXiv:1901.02860}.

\bibitem[{Devlin et~al.(2018)Devlin, Chang, Lee, and
  Toutanova}]{devlin2018bert}
Devlin, J.; Chang, M.-W.; Lee, K.; and Toutanova, K. 2018.
\newblock Bert: Pre-training of deep bidirectional transformers for language
  understanding.
\newblock \emph{arXiv preprint arXiv:1810.04805}.

\bibitem[{Ding et~al.(2020)Ding, Zhou, Yang, and Tang}]{ding2020cogltx}
Ding, M.; Zhou, C.; Yang, H.; and Tang, J. 2020.
\newblock Cogltx: Applying bert to long texts.
\newblock \emph{Advances in Neural Information Processing Systems}, 33:
  12792--12804.

\bibitem[{He et~al.(2019)He, Wang, Liu, Feng, and Wu}]{he2019long}
He, J.; Wang, L.; Liu, L.; Feng, J.; and Wu, H. 2019.
\newblock Long document classification from local word glimpses via recurrent
  attention learning.
\newblock \emph{IEEE Access}, 7: 40707--40718.

\bibitem[{Huang et~al.(2021)Huang, Tao, Huang, Xiong, and
  Yu}]{huang2021hierarchical}
Huang, W.; Tao, Z.; Huang, X.; Xiong, L.; and Yu, J. 2021.
\newblock Hierarchical Self-Attention Hybrid Sparse Networks for Document
  Classification.
\newblock \emph{Mathematical Problems in Engineering}, 2021.

\bibitem[{Pappagari et~al.(2019)Pappagari, Zelasko, Villalba, Carmiel, and
  Dehak}]{pappagari2019hierarchical}
Pappagari, R.; Zelasko, P.; Villalba, J.; Carmiel, Y.; and Dehak, N. 2019.
\newblock Hierarchical transformers for long document classification.
\newblock In \emph{2019 IEEE Automatic Speech Recognition and Understanding
  Workshop (ASRU)}, 838--844. IEEE.

\bibitem[{Park, Vyas, and Shah(2022)}]{park2022efficient}
Park, H.~H.; Vyas, Y.; and Shah, K. 2022.
\newblock Efficient Classification of Long Documents Using Transformers.
\newblock \emph{arXiv preprint arXiv:2203.11258}.

\bibitem[{Rae et~al.(2019)Rae, Potapenko, Jayakumar, and
  Lillicrap}]{rae2019compressive}
Rae, J.~W.; Potapenko, A.; Jayakumar, S.~M.; and Lillicrap, T.~P. 2019.
\newblock Compressive transformers for long-range sequence modelling.
\newblock \emph{arXiv preprint arXiv:1911.05507}.

\bibitem[{Si and Roberts(2021)}]{si2021hierarchical}
Si, Y.; and Roberts, K. 2021.
\newblock Hierarchical transformer networks for longitudinal clinical document
  classification.
\newblock \emph{arXiv preprint arXiv:2104.08444}.

\bibitem[{Sukhbaatar et~al.(2019)Sukhbaatar, Grave, Bojanowski, and
  Joulin}]{sukhbaatar2019adaptive}
Sukhbaatar, S.; Grave, E.; Bojanowski, P.; and Joulin, A. 2019.
\newblock Adaptive attention span in transformers.
\newblock \emph{arXiv preprint arXiv:1905.07799}.

\bibitem[{Sun et~al.(2019)Sun, Qiu, Xu, and Huang}]{sun2019fine}
Sun, C.; Qiu, X.; Xu, Y.; and Huang, X. 2019.
\newblock How to fine-tune bert for text classification?
\newblock In \emph{China national conference on Chinese computational
  linguistics}, 194--206. Springer.

\bibitem[{Tay et~al.(2020{\natexlab{a}})Tay, Dehghani, Abnar, Shen, Bahri,
  Pham, Rao, Yang, Ruder, and Metzler}]{tay2020long}
Tay, Y.; Dehghani, M.; Abnar, S.; Shen, Y.; Bahri, D.; Pham, P.; Rao, J.; Yang,
  L.; Ruder, S.; and Metzler, D. 2020{\natexlab{a}}.
\newblock Long range arena: A benchmark for efficient transformers.
\newblock \emph{arXiv preprint arXiv:2011.04006}.

\bibitem[{Tay et~al.(2020{\natexlab{b}})Tay, Dehghani, Bahri, and
  Metzler}]{tay2020efficient}
Tay, Y.; Dehghani, M.; Bahri, D.; and Metzler, D. 2020{\natexlab{b}}.
\newblock Efficient transformers: A survey.
\newblock \emph{ACM Computing Surveys (CSUR)}.

\bibitem[{Wagh et~al.(2021)Wagh, Khandve, Joshi, Wani, Kale, and
  Joshi}]{wagh2021comparative}
Wagh, V.; Khandve, S.; Joshi, I.; Wani, A.; Kale, G.; and Joshi, R. 2021.
\newblock Comparative study of long document classification.
\newblock In \emph{TENCON 2021-2021 IEEE Region 10 Conference (TENCON)},
  732--737. IEEE.

\bibitem[{Wang et~al.(2019)Wang, Ng, Ma, Nallapati, and Xiang}]{wang2019multi}
Wang, Z.; Ng, P.; Ma, X.; Nallapati, R.; and Xiang, B. 2019.
\newblock Multi-passage bert: A globally normalized bert model for open-domain
  question answering.
\newblock \emph{arXiv preprint arXiv:1908.08167}.

\bibitem[{Wang et~al.(2020)Wang, Wang, Zhang, Duan, Zhou, and
  Chen}]{wang2020learning}
Wang, Z.; Wang, C.; Zhang, H.; Duan, Z.; Zhou, M.; and Chen, B. 2020.
\newblock Learning dynamic hierarchical topic graph with graph convolutional
  network for document classification.
\newblock In \emph{International Conference on Artificial Intelligence and
  Statistics}, 3959--3969. PMLR.

\end{thebibliography}

\section{Acknowledgments}
Author is especially grateful to the Research Department in Elm Company for funding and support this project.

\end{document}